\documentclass[conference]{IEEEtran}
\IEEEoverridecommandlockouts

\usepackage{cite}
\usepackage{amsmath,amssymb,amsfonts}
\usepackage{algorithmic}
\usepackage{graphicx}
\usepackage{textcomp}
\usepackage[table]{xcolor}
\usepackage{subcaption}
\usepackage{algorithm2e}
\usepackage{hyperref}
\usepackage{amsfonts}
\usepackage{booktabs}
\usepackage{array}
\usepackage{siunitx}
\usepackage{multirow}
\usepackage{url}
\usepackage{pifont}

\newcommand{\cmark}{\ding{51}}%
\newcommand{\xmark}{\ding{55}}%
\newcolumntype{P}[1]{>{\centering\arraybackslash}p{#1}}
\newcolumntype{M}[1]{>{\centering\arraybackslash}m{#1}}

\def\BibTeX{{\rm B\kern-.05em{\sc i\kern-.025em b}\kern-.08em
    T\kern-.1667em\lower.7ex\hbox{E}\kern-.125emX}}
\begin{document}

\title{Personalized Behaviour Models: A Survey Focusing on Autism Therapy Applications
}

\author{\IEEEauthorblockN{ Michał Stolarz}
\IEEEauthorblockA{\textit{Hochschule Bonn-Rhein-Sieg} \\
Sankt Augustin, Germany \\
michal.stolarz@smail.inf.h-brs.de}
\and
\IEEEauthorblockN{ Alex Mitrevski}
\IEEEauthorblockA{\textit{Hochschule Bonn-Rhein-Sieg} \\
Sankt Augustin, Germany \\
aleksandar.mitrevski@h-brs.de}
\and
\IEEEauthorblockN{ Mohammad Wasil}
\IEEEauthorblockA{\textit{Hochschule Bonn-Rhein-Sieg} \\
Sankt Augustin, Germany \\
mohammad.wasil@h-brs.de}
\and
\IEEEauthorblockN{ Paul G. Plöger}
\IEEEauthorblockA{\textit{Hochschule Bonn-Rhein-Sieg} \\
Sankt Augustin, Germany \\
paul.ploeger@h-brs.de}
}

\maketitle

\begin{abstract}
Children with Autism Spectrum Disorder find robots easier to communicate with than humans. Thus, robots have been introduced in autism therapies. However, due to the environmental complexity, the used robots often have to be controlled manually. This is a significant drawback of such systems and it is required to make them more autonomous. In particular, the robot should interpret the child's state and continuously adapt its actions according to the behaviour of the child under therapy. This survey elaborates on different forms of personalized robot behaviour models. Various approaches from the field of Human-Robot Interaction, as well as Child-Robot Interaction, are discussed. The aim is to compare them in terms of their deficits, feasibility in real scenarios, and potential usability for autism-specific Robot-Assisted Therapy. The general challenge for algorithms based on which the robot learns proper interaction strategies during therapeutic games is to increase the robot's autonomy, thereby providing a basis for a robot's decision-making.

\end{abstract}


\section{Introduction}
\label{sec:introduction}
In the European Union, there are over 5 million people affected by autism~\cite{deenigma2022} and it is estimated that 1 in 160 children all over the world is diagnosed with Autism Spectrum Disorder (ASD)~\cite{jain2020modeling}. People with ASD often have difficulties in social interaction and communication. To alleviate the effects of ASD, individualized therapies are provided. However, autistic children find robots easier to communicate with than humans~\cite{robins2006}, thus Robot-Assisted Therapies (RATs) have been being investigated. During RAT, most of the time therapists have to control the robot remotely (Wizard of Oz approach)~\cite{deenigma2022}~\cite{david2018developing,robins2017developing,rudovic2017measuring,marinoiu20183d}. Because of it, the therapist might not be able to fully focus on the therapy and react appropriately to the child's behaviour~\cite{cao2018personalized}. To reduce their workload, the autonomy of the robot has to be increased, namely it should be able to interpret a child’s behaviour and adapt its actions to the individual needs of the child~\cite{esteban2017build}.

\begin{figure}
    \begin{subfigure}[b]{0.24\textwidth}
        \includegraphics[width=\textwidth]{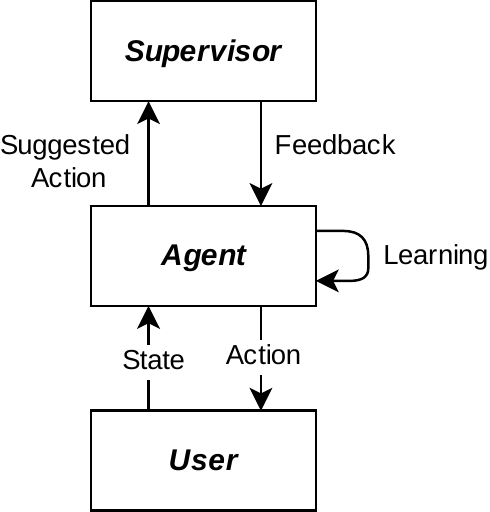}
        \subcaption{}%
        \label{fig:guidance_learning}
    \end{subfigure}
    \begin{subfigure}[b]{0.24\textwidth}
        \includegraphics[width=0.73\textwidth]{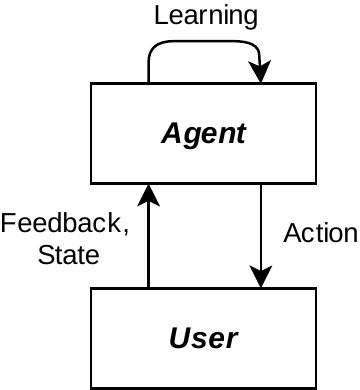}
        \subcaption{}%
        \label{fig:feedback_learning}
    \end{subfigure}

\caption{Two methods of agent interactive learning: (a) learning from guidance, (b) learning from feedback}
\label{fig:adaptive_system_learning}
\end{figure}

Adaptation is possible if the robot actively learns a user model that encodes certain attributes of the user. The user model can be integrated into a robot decision-making algorithm~\cite{rossi2017user} called a behaviour model, which allows the system to choose appropriate robot reactions in response to the actions of each individual user. Personalization refers to the adaptation of the system to the individual user over time~\cite{rossi2017user} and can be solved by using Interactive Machine Learning (IML), which involves the user in the learning loop~\cite{senft2019teaching}. IML usually makes use of \emph{learning from guidance} or \emph{learning from feedback}. Learning from guidance relies on an external supervisor (e.g. therapist), who provides expert knowledge to the system (\autoref{fig:guidance_learning}). The supervisor is able to assess the decisions of the robot before being executed, namely they are able to accept, or alternatively reject and override the suggested reaction of the robot. This solution guarantees that the system will not execute any undesirable actions during learning, but is sensitive to the mistakes of the supervising person. On the other hand, learning from feedback uses direct feedback from the user (e.g. engagement level of the user) (\autoref{fig:feedback_learning}). As there is no supervising person, the robot has to explore by itself what effects its actions have.

In this work, different personalized behaviour models present in the Human-Robot Interaction (HRI) field are described and compared. The discussed concepts are based on the aforementioned IML techniques and are compared in terms of their potential usability and feasibility in real-life autism RATs. We believe that the provided survey can aid the design of solutions for increasing the autonomy of robots used in ASD therapy. Section~\ref{sec:related_work} describes other surveys related to the topic of adaptation and personalization techniques in HRI. Section~\ref{sec:literature_review} provides an analysis of four personalization techniques and elaborates on their usefulness in autism therapies. Various open challenges are presented in section~\ref{sec:challenges}. Section \ref{sec:conclusion} provides a conclusion and elaborates on our planned future work.

\section{Related Surveys}
\label{sec:related_work}
The related literature provides a structured taxonomy of robotic systems capable of adapting to user differences. Martins et al.~\cite{martins2019user} proposed to categorize user-adaptive systems into three classes: (i) systems with no user model (dominated by reactive behaviour, no user information is maintained), (ii) systems with a static user model (information about the user is used, but defined apriori), and (iii) systems based on dynamic user models (the information about the user is continuously updated through interaction). In~\cite{nocentini2019survey}, a robot's adaptiveness to a user is specified as behavioural adaptation and several approaches with and without experimentation setups are described. Rossi et al.~\cite{rossi2017user} elaborate on different forms of behaviour adaptation that are possible when developing a social robot, namely physical, cognitive and social adaptation. In this related literature, there is a limited elaboration on methods of providing robot adaptation in healthcare and education applications (which is the main interest of our work), i.e. improving the user's performance in certain tasks.

These topics are addressed in~\cite{clabaugh2019escaping} and~\cite{tsiakas2018taxonomy}. Tsiakas et al.~\cite{tsiakas2018taxonomy} provide a generic taxonomy based on the following categories: task type and requirements, interaction types and roles (of the robot and humans in the interaction), level of autonomy, personalization dimensions (e.g. task difficulty, supportive feedback, proxemics). The healthcare domain was especially covered in~\cite{cao2017survey}, where robot behaviour control architectures are classified based on the robot's role in the interaction, namely: companion (assisting humans in carrying out a variety of tasks), therapeutic play partner or peer (used to support human therapists in therapeutic games) and coach (conducting therapeutic scenarios by providing task descriptions, feedback, and monitoring the user's performance). Although several approaches for children with ASD are mentioned in~\cite{cao2017survey}, many do not address the problem of user adaptation. Kubota and Riek~\cite{kubota2021methods} elaborate on behaviour adaptation in robotics applied to neurorehabilitation and describe commonly used algorithms in that field as well. While the above literature mentions adaptation methodologies used in therapies with children with ASD, there is a lack of comparisons of personalization behaviour models from different HRI domains in order to find approaches that will help in designing solutions for RATs for children with ASD.

\section{Review of Personalized Behaviour Models}
\label{sec:literature_review}
This work provides an overview of four personalization problems that are solved in the context of HRI. The found approaches are analyzed in terms of the usability for RAT for children with ASD. They are categorized into four personalization dimensions~\cite{tsiakas2018taxonomy} (here we consider only aspects in which the robot control can be adjusted): social behaviour, game difficulty, affection, and user preferences (e.g. proxemics). It is particularly desired that the robot is able to personalize its behaviour such that it improves a child’s performance in therapeutic games (with or without the help of a therapist). This can be done by adjusting the difficulty of the games to each individual child. Additionally, the robot should react properly when the interaction with a child does not go as planned, which means that it should prevent the child from getting bored, disengaged or demotivated, for instance by providing reengaging and motivating feedback.

\subsection{Literature inclusion criteria}
To find relevant publications, we used the Google Scholar search engine. We performed a search based on the following keywords: \emph{robot personalization}, \emph{robot adaptation}, \emph{Robot-Assisted Therapy}, \emph{autism}, \emph{Autism Spectrum Disorder}, \emph{behaviour model}. In a first stage of the search, we looked for publications that relate to the concept of continuous adaptation in HRI. Secondly, we limited our target field to domains closer to ASD therapy, namely robots as learning tutors or peers, such that we included existing approaches directly associated with robot ASD therapy as well.

\subsection{Social behaviour personalization}
Social behaviour personalization refers to how a robot adapts its gestures, facial expressions, and language content (e.g. type of feedback) to a user. The aim of this personalization technique is to maintain user involvement in the interaction. One of the state-of-the-art solutions is the supervised autonomy system for RAT for children with ASD, which was introduced in the DREAM project~\cite{esteban2017build,cao2019robot}. In this system, the robot produces actions according to therapeutic scripts defined by therapists; however, when the interaction does not go as planned, the robot tries to seek appropriate actions on its own~\cite{cao2019robot}. Here, supervised autonomy means that the robot is able to adapt its behaviour to the behaviour of the children (after assessing it based on sensor data). This is similar to the learning from guidance concept (\autoref{fig:guidance_learning}) as, before executing any actions, the robot requests the therapist for feedback about its suitability. Once feedback has been obtained (the therapist accepts or overrides the suggested action), the robot executes the appropriate action and includes the feedback into its behavioural model for learning. Then, the new state of the child is acquired. This approach has been successfully deployed and evaluated in real-world scenarios with children with ASD~\cite{cao2019robot}, but it was not personalized, as the learning procedure was performed on data from all children that were participating in the experiments.

One proposed algorithm for decision-making is a feed-forward network~\cite{senft2015sparc}. This has good generalisation abilities and can be personalized to a specific person~\cite{senft2015human}, but has to be retrained every time therapist feedback is obtained. This might make this solution inappropriate for online real-time interactions in case of long-time scenarios~\cite{senft2018teaching}, as the learning time increases with the amount of collected data.
Another proposed solution is based on reinforcement learning with the Q-learning algorithm~\cite{watkins1992q,senft2017supervised}; however, in Q-learning, a considerable amount of data is needed to obtain an optimal policy, which means that a significant number of interactions with the user is required. To guarantee fast convergence of this algorithm, the problem has to be decomposed so that the Q-value table stays relatively small~\cite{hemminahaus2017towards}. To reduce the memory requirements and make the learning converge faster, the MAXQ hierarchical reinforcement learning algorithm~\cite{dietterich2000hierarchical} is used in~\cite{chan2012social}, where a robot providing personalized assistive behaviours for a memory game is developed. This memory deficit was faced in~\cite{senft2019teaching} as well, where applying nearest neighbors allowed to obtain a reasonable training time; this result was achieved in spite of the high dimensionality of the state and action spaces. The aforementioned approach was tested in a study with 75 typically developing children playing an educational game about food webs.

Most of the aforementioned approaches are based on learning from guidance~\cite{esteban2017build,senft2019teaching,senft2015sparc,senft2017supervised}, which is advantageous especially for systems where robot mistakes imply ethical concerns, but is very dependent on the supervisor. This can have a negative impact on the learned policy, especially when the supervisor makes wrong decisions. The effects of therapist mistakes can be alleviated if the reinforcement signal is obtained directly from the user. This is done by applying learning from feedback (\autoref{fig:feedback_learning}), where the robot has to find appropriate actions on its own. The concept of combining learning from guidance and feedback is also presented in~\cite{tsiakas2016adaptive}. In~\cite{leyzberg2014personalizing,hemminahaus2017towards,chan2012social} approaches for personalizing a robot's behaviour in real-life scenario based exclusively on learning from feedback are described. In~\cite{leyzberg2014personalizing}, the robot was deployed in the role of a tutor that was giving lessons when a user was playing a nonogram puzzle game. This approach is limited to providing users only with lessons that complement their missing knowledge and is not able to react when the interaction does not go as planned (e.g. the user becomes disengaged in the game).

\subsection{Game difficulty personalization}

The solutions presented so far are adaptive in terms of the reactions to a user's behaviour. However, it is also important to autonomously adapt the game that is the basis of the interaction during a therapy. Particularly in RAT for children with ASD, there is a need for game difficulty personalization that would match each child's skill level. Systems that can provide the aforementioned adaptation are based mainly on learning from feedback. A personalization concept based on adapting the progression of a lesson to a user's performance is covered in~\cite{jain2020modeling,clabaugh2019long,baxter2017robot,scassellati2018improving}. In~\cite{jain2020modeling,clabaugh2019long} reinforcement learning is used in order to personalize feedback and instruction difficulty levels during math games. In~\cite{baxter2017robot} Baxter et al. deploy a rule-based adaptation algorithm, for instance based on a comparison of the number of successfully completed tasks to a predefined threshold. Here, only experiments with typically developing children were conducted. Moreover, in~\cite{jain2020modeling,clabaugh2019long,scassellati2018improving}, a learning-based decision-making algorithm that would generate proper actions in case of deviations of a user’s behaviour (e.g. the child is disengaged or demotivated) is not used.

This deficit is faced in~\cite{tsiakas2018task}, where Q-learning is used so that a robot can adapt the difficulty of a game as well as provide encouraging or challenging feedback to the user. In addition, an investigation on how different methods of updating the Q-table can increase the speed of convergence of the policy is presented and user models that can be used for initial training or testing of the proposed framework are provided.
Here, it is suggested not to start learning a policy from scratch (in case there is a new user of the system), but to start from the policy already learned during tests with a certain user model. This concept is applied in~\cite{tsiakas2016adaptive}, where it is shown that it indeed reduces the number of iterations required for the policy to converge. It should, however, be mentioned that neither~\cite{tsiakas2018task} nor~\cite{tsiakas2016adaptive} presented experiments with real users and in both cases, the created user models do not reflect the behaviour of a person with ASD.

\subsection{Affective personalization}
\label{subsec:affective-personalization}
Affective personalization is another aspect of personalization of a robot's behaviour that is based on learning from feedback. Here, the robot is supposed to explicitly maintain an internal affective state, which impacts its interaction with the user. In~\cite{cao2018personalized,gordon2016affective}, the effect of the robot acting as a personal character that has its own affect (e.g. emotion) is investigated. In particular, the robot's affect is influencing the robot's non-verbal and/or verbal behaviour during the human-robot interaction. In~\cite{gordon2016affective}, using the SARSA reinforcement learning algorithm~\cite{singh1996reinforcement}, the robot is learning how to maximise the engagement and valence (positiveness of emotion) of a child during a game. Here, the action space represents the change in engagement and valence of the robot. The policy was however, not able to converge during the interaction, which was performed over seven sessions. In~\cite{cao2018personalized}, the affective state of a robot is influenced by an occurred event and the robot's personality according to the Orthony Claire Collins model~\cite{ortony1990cognitive}. However, if and how the personality of the robot was adapted to the child's one is not mentioned.

A robot having its own affect has not been shown to be advantageous during autism therapy, as it is suspected that affective adaptation may cause children with ASD to become overwhelmed~\cite{cao2018personalized}. Moreover, autistic children usually have difficulties recognizing and expressing emotions~\cite{rudovic2017measuring}; thus, a system that estimates the child's affective state and uses it for learning may lead to a suboptimal behaviour model.

\subsection{Personalization based on user’s preferences}
\label{subsec:preferences-personalization}
A robotic system may also be able to create a model of the preferences of a user~\cite{martins2017bum,martins2018bum}, such as the preferred speech volume of the robot or the robot's distance to the user. Having that information, the robot can adapt its actions to the user's individual needs~\cite{martins2019alphapomdp,karami2016adaptive}. This field of user-adaptive robotics is based on the concept of learning from feedback and, to the best of our knowledge, is not covered in RATs for children with ASD. However, as such children often have sensory difficulties~\cite{javed2019robotic}, incorporating a model of their preferences in the robot's decision-making process may be advantageous. The contribution of~\cite{martins2017bum,martins2018bum} is an approach for creating user models, but no decision-making algorithms are provided. Additionally, experiments are mainly conducted in simulation or with a relatively small number of real users.

In~\cite{martins2019alphapomdp,karami2016adaptive}, robots are able to model user preferences and use the collected knowledge to choose appropriate actions. These solutions are based on an inverse reinforcement learning~\cite{karami2016adaptive} and a model-based reinforcement learning~\cite{martins2019alphapomdp}.
The main disadvantage of~\cite{karami2016adaptive} is the fact that the communication with the robot is limited as it can recognize only four hand movements; however, an algorithm that enhances the learning speed of the robot by identifying only relevant variables that define the robot's state is provided. In~\cite{martins2019alphapomdp}, an approach where a robot is rewarded for its actions according to their impact is presented, such that the robot’s knowledge about the impact of the actions is continuously updated. The proposed algorithm was tested in real and simulated trials, but was not able to converge during tests with human users.

\section{Challenges}
\label{sec:challenges}
After the analysis of different personalization approaches, a few challenges in terms of their application in the field of RAT for children with ASD remain. These challenges are related to the fact that the developed system has to learn and react quickly so that it can be used in real-life interventions. The four identified challenges are as follows:

\begin{itemize}
\item maintenance of an adequately big state and action space enabling the robot to personalize the game difficulty and its reactions to a specific child’s behaviour (e.g. disengagement during the game),
\item fast convergence to an optimal policy without the need for a significant number of interactions with the user, or finding a policy that is sufficient for effective practical interaction,
\item evaluation during interaction with real people, and
\item resistance to the supervising person’s mistakes.
\end{itemize}

While this survey focuses on RAT for children with ASD, it should be noted that first three challenges are independent of this domain and could be considered common for diverse HRI applications. The first challenge consists of two aspects. First of all, the developed model should help therapists in improving a child’s performance; this can be done by adjusting the difficulty of games according to each child’s skill level. Secondly, the robot should prevent the user from getting bored or demotivated during therapy, by providing, for example, motivating feedback. In many approaches, only one of those aspects was covered. Additionally, the small size of the action/state space often significantly reduces the capabilities of the robot. These problems are present in~\cite{jain2020modeling,senft2017supervised,gordon2016affective,scassellati2018improving,tsiakas2016adaptive,senft2015sparc,clabaugh2019long,leyzberg2014personalizing}. The second challenge is related to the need for the system to quickly adapt to the user, for example in order to reduce the workload of the supervisor as quickly as possible. That aspect was a deficit especially in~\cite{senft2017supervised,gordon2016affective,martins2019alphapomdp,clabaugh2019long}. The third challenge is related to the fact that many approaches have not been tested in human trials. As it is usually difficult to conduct such studies, many proposed personalization methods are tested with user simulations; for instance, the personalization algorithms in~\cite{senft2017supervised,tsiakas2016adaptive,senft2015sparc,tsiakas2018task} were not tested on robots interacting with human users. The last challenge is relevant mainly for approaches that are based on learning from guidance, namely inappropriate decisions of the person supervising the robot may significantly affect the approaches presented in~\cite{esteban2017build,cao2019robot,senft2017supervised,senft2019teaching,senft2015sparc}.

The first challenge could be addressed by applying the algorithm presented by Senft et al.~\cite{senft2019teaching}. A potential solution for the second and fourth challenge would be to combine the learning from feedback and learning from guidance approaches (as done in~\cite{tsiakas2016intelligent,tsiakas2016adaptive}), which would make the system more resistant to the supervisor's mistakes and would improve the speed of policy convergence~\cite{tsiakas2016adaptive}. The latter can be also achieved by starting the learning process from a generic policy (pretrained on user simulations~\cite{tsiakas2018task}) rather than from scratch. Moreover, this method could help in addressing the problem of generalization between users while accounting for individual differences.

\section{Conclusion}
\label{sec:conclusion}

This work aims at providing an overview of different personalization techniques and a discussion on their applicability in the therapy for children with ASD. Approaches in four different personalization dimensions were discussed; however, based on the surveyed literature, we conclude that two techniques are mostly used and are adequate for autism-specific RAT, namely social behaviour personalization and game difficulty personalization. We also identified the deficits of various recent approaches and formulated the challenges in applying them in the context of autism therapy.

As future work, we plan to develop a robot that would support the therapist when conducting therapy for children with ASD. We particularly want to address the first two of the aforementioned challenges, namely the robot will autonomously personalize the therapy game content to the child’s individual skills. To maximise the learning gains, the robot will also react if a child’s disengagement or demotivation is perceived. Another requirement is that the personalized behaviour model should converge fast so that no significant amount of interactions with the child is necessary; for this, we are considering the use of active learning and policy pretraining.

\section*{Acknowledgment}
This work is conducted in the context of the MigrAVE project funded by the German Ministry of Education and Research (BMBF). We hereby thank our partners, Münster University of Applied Sciences (FHM) and the RFH – University of Applied Sciences, Cologne.

\appendix
A comparison of the analyzed behaviour models is provided in Table~\ref{tab:models_comparison}. Each row describes one of the aforementioned approaches and contains a reference, the aspect in which the robot was personalized, and the type of personalization learning algorithm (learning from feedback or learning from guidance). The name of the personalization algorithm and features that were used for learning are specified as well. Additionally, the last column contains information on whether the behaviour model was tested in long-term operation. This means that the system repeatedly interacted with the same user over a period of time that was longer than one week.

\begin{table*}[htp]
 \caption{Comparison of the personalized behaviour models}
\label{tab:models_comparison}
\centering
\begin{tabular}{M{1.5cm}M{2.5cm}M{2.5cm}M{2.5cm}M{2.5cm}M{2.5cm}} \toprule
    \cellcolor{gray!10!white} \textbf{Reference} & \cellcolor{gray!10!white} \textbf{Personalization aspect} & \cellcolor{gray!10!white} \textbf {Type of learning} & \cellcolor{gray!10!white} \textbf{Personalization algorithm} & \cellcolor{gray!10!white} \textbf{Features used for learning} & \cellcolor{gray!10!white} \textbf{Demonstrated long-term operation ($>$1 week)}\\ \midrule
     \cite{senft2015sparc,senft2015human}          & \multirow{12}{*}{\hfil Social behaviour}  & Guidance & Feed-forward network  & User's engagement, motivation, game progress & \xmark \\
    \cite{senft2017supervised}      &                                    & Guidance & Q-learning  & Characteristics of the simulation environment & \xmark\\
    \cite{senft2019teaching}        &                                    & Guidance & Nearest neighbors & Game progress and interaction state &  \xmark \\
    \cite{hemminahaus2017towards}   &                                    & Feedback & Q-learning  & Gaze and speech behavior of the user, game progress & \xmark \\
    \cite{chan2012social}           &                                    & Feedback & MAXQ  & User's arosal and speech, game progress  & \xmark\\
    \cite{leyzberg2014personalizing}&                                    & Feedback & Bayessian network  & User's progress in skill learning & \xmark\\ \hline
    \cite{tsiakas2016adaptive}      & \multirow{6}{*}{Task difficulty}  & Feedback, guidance & Q-learning  & Game progress & \xmark\\
    \cite{jain2020modeling,clabaugh2019long} &                          & Feedback & Q-learning  & Game progress, number of user's help requests & \cmark\\
    \cite{baxter2017robot}          &                                   & -- & Rule-based & Game progress & \cmark \\
    \cite{scassellati2018improving} &                                   & -- & -- & Game progress & \cmark \\ \hline
    \cite{tsiakas2018task} & Social behaviour, task difficulty & Feedback & Q-learning  & User's engagement and game progress & \xmark \\ \hline
    \cite{gordon2016affective} & Affection  & Feedback & SARSA & User's engagement, valence, game performance & \cmark\\ \hline
    \cite{karami2016adaptive} & \multirow{6}{*}{User's preferences}  & Feedback & Inverse reinforcement learning algorithm  & User's age, gender and activity preferences; environmental noise and brightness; daytime  & \xmark \\
    \cite{martins2019alphapomdp} &                              & Feedback & Model-based reinforcement learning algorithm  & User's satisfaction and health levels, robot's speaking volume and distance to the user, time of day& \xmark\\ \bottomrule
\end{tabular}
\end{table*}

\bibliographystyle{ieeetr}
\bibliography{bibliography.bib}

\begin{thebibliography}{10}

\bibitem{deenigma2022}
``{DE-ENIGMA} {P}layfully {E}mpowering {A}utistic {C}hildren.''
  \url{https://de-enigma.eu/background-of-the-project/}.
\newblock Accessed: 2022-02-05.

\bibitem{jain2020modeling}
S.~Jain, B.~Thiagarajan, Z.~Shi, C.~Clabaugh, and M.~J. Matari{\'c}, ``Modeling
  engagement in long-term, in-home socially assistive robot interventions for
  children with autism spectrum disorders,'' {\em Science Robotics}, vol.~5,
  no.~39, 2020.

\bibitem{robins2006}
B.~Robins, K.~Dautenhahn, and J.~Dubowski, ``{Does appearance matter in the
  interaction of children with autism with a humanoid robot?},'' {\em
  Interaction Studies}, vol.~7, no.~3, pp.~479--512, 2006.

\bibitem{david2018developing}
D.~O. David, C.~A. Costescu, S.~Matu, A.~Szentagotai, and A.~Dobrean,
  ``Developing joint attention for children with autism in robot-enhanced
  therapy,'' {\em Int. Journal of Social Robotics}, vol.~10, no.~5,
  pp.~595--605, 2018.

\bibitem{robins2017developing}
B.~Robins, K.~Dautenhahn, L.~Wood, and A.~Zaraki, ``Developing interaction
  scenarios with a humanoid robot to encourage visual perspective taking skills
  in children with autism--preliminary proof of concept tests,'' in {\em Int.
  Conf. on Social Robotics}, pp.~147--155, Springer, 2017.

\bibitem{rudovic2017measuring}
O.~Rudovic, J.~Lee, L.~Mascarell-Maricic, B.~W. Schuller, and R.~W. Picard,
  ``Measuring engagement in robot-assisted autism therapy: a cross-cultural
  study,'' {\em Frontiers in Robotics and AI}, vol.~4, p.~36, 2017.

\bibitem{marinoiu20183d}
E.~Marinoiu, M.~Zanfir, V.~Olaru, and C.~Sminchisescu, ``3d human sensing,
  action and emotion recognition in robot assisted therapy of children with
  autism,'' in {\em Proc. of the IEEE Conf. on Computer Vision and Pattern
  Recognition}, pp.~2158--2167, 2018.

\bibitem{cao2018personalized}
H.-L. Cao {\em et~al.}, ``A personalized and platform-independent behavior
  control system for social robots in therapy: development and applications,''
  {\em IEEE Trans. Cognitive and Developmental Systems}, vol.~11, no.~3,
  pp.~334--346, 2018.

\bibitem{esteban2017build}
P.~G. Esteban {\em et~al.}, ``How to build a supervised autonomous system for
  robot-enhanced therapy for children with autism spectrum disorder,'' {\em
  Paladyn, Journal of Behavioral Robotics}, vol.~8, no.~1, pp.~18--38, 2017.

\bibitem{rossi2017user}
S.~Rossi, F.~Ferland, and A.~Tapus, ``User profiling and behavioral adaptation
  for {HRI}: {A} survey,'' {\em Pattern Recognition Letters}, vol.~99,
  pp.~3--12, 2017.

\bibitem{senft2019teaching}
E.~Senft, S.~Lemaignan, P.~E. Baxter, M.~Bartlett, and T.~Belpaeme, ``Teaching
  robots social autonomy from in situ human guidance,'' {\em Science Robotics},
  vol.~4, no.~35, 2019.

\bibitem{martins2019user}
G.~S. Martins, L.~Santos, and J.~Dias, ``User-adaptive interaction in social
  robots: {A} survey focusing on non-physical interaction,'' {\em Int. Journal
  of Social Robotics}, vol.~11, no.~1, pp.~185--205, 2019.

\bibitem{nocentini2019survey}
O.~Nocentini {\em et~al.}, ``A survey of behavioral models for social robots,''
  {\em Robotics}, vol.~8, no.~3, p.~54, 2019.

\bibitem{clabaugh2019escaping}
C.~Clabaugh and M.~Matari{\'c}, ``Escaping oz: Autonomy in socially assistive
  robotics,'' {\em Annual Review of Control, Robotics, and Autonomous Systems},
  vol.~2, pp.~33--61, 2019.

\bibitem{tsiakas2018taxonomy}
K.~Tsiakas, M.~Kyrarini, V.~Karkaletsis, F.~Makedon, and O.~Korn, ``A taxonomy
  in robot-assisted training: current trends, needs and challenges,'' {\em
  Technologies}, vol.~6, no.~4, p.~119, 2018.

\bibitem{cao2017survey}
H.-L. Cao {\em et~al.}, ``A survey on behavior control architectures for social
  robots in healthcare interventions,'' {\em Int. Journal of Humanoid
  Robotics}, vol.~14, no.~04, p.~1750021, 2017.

\bibitem{kubota2021methods}
A.~Kubota and L.~D. Riek, ``Methods for robot behavior adaptation for cognitive
  neurorehabilitation,'' {\em Annual Review of Control, Robotics, and
  Autonomous Systems}, vol.~5, 2021.

\bibitem{cao2019robot}
H.-L. Cao {\em et~al.}, ``Robot-enhanced therapy: {D}evelopment and validation
  of supervised autonomous robotic system for autism spectrum disorders
  therapy,'' {\em IEEE {R}obotics \& {A}utomation {M}agazine}, vol.~26, no.~2,
  pp.~49--58, 2019.

\bibitem{senft2015sparc}
E.~Senft, P.~Baxter, J.~Kennedy, and T.~Belpaeme, ``Sparc: Supervised
  progressively autonomous robot competencies,'' in {\em Int. Conf. Social
  Robotics {ICSR}}, pp.~603--612, Springer, 2015.

\bibitem{senft2015human}
E.~Senft, P.~Baxter, and T.~Belpaeme, ``Human-guided learning of social action
  selection for robot-assisted therapy,'' in {\em Machine Learning for
  Interactive Systems}, pp.~15--20, PMLR, 2015.

\bibitem{senft2018teaching}
E.~Senft, {\em {T}eaching {R}obots {S}ocial {A}utonomy {F}rom {I}n {S}itu
  {H}uman {S}upervision}.
\newblock PhD thesis, University of Plymouth, 2018.

\bibitem{watkins1992q}
C.~J. Watkins and P.~Dayan, ``Q-learning,'' {\em Machine learning}, vol.~8,
  no.~3-4, pp.~279--292, 1992.

\bibitem{senft2017supervised}
E.~Senft, P.~Baxter, J.~Kennedy, S.~Lemaignan, and T.~Belpaeme, ``Supervised
  autonomy for online learning in human-robot interaction,'' {\em Pattern
  Recognition Letters}, vol.~99, pp.~77--86, 2017.

\bibitem{hemminahaus2017towards}
J.~Hemminahaus and S.~Kopp, ``Towards adaptive social behavior generation for
  assistive robots using reinforcement learning,'' in {\em 2017 12th ACM/IEEE
  Int. Conf. on Human-Robot Interaction (HRI)}, pp.~332--340, IEEE, 2017.

\bibitem{dietterich2000hierarchical}
T.~G. Dietterich, ``Hierarchical reinforcement learning with the maxq value
  function decomposition,'' {\em Journal of Artificial Intelligence Research},
  vol.~13, pp.~227--303, 2000.

\bibitem{chan2012social}
J.~Chan and G.~Nejat, ``Social intelligence for a robot engaging people in
  cognitive training activities,'' {\em Int. Journal of Advanced Robotic
  Systems}, vol.~9, no.~4, p.~113, 2012.

\bibitem{tsiakas2016adaptive}
K.~Tsiakas, M.~Dagioglou, V.~Karkaletsis, and F.~Makedon, ``Adaptive robot
  assisted therapy using interactive reinforcement learning,'' in {\em Int.
  Conf. on Social Robotics}, pp.~11--21, Springer, 2016.

\bibitem{leyzberg2014personalizing}
D.~Leyzberg, S.~Spaulding, and B.~Scassellati, ``Personalizing robot tutors to
  individuals’ learning differences,'' in {\em 2014 9th ACM/IEEE Int. Conf.
  on Human-Robot Interaction (HRI)}, pp.~423--430, IEEE, 2014.

\bibitem{clabaugh2019long}
C.~Clabaugh {\em et~al.}, ``Long-term personalization of an in-home socially
  assistive robot for children with autism spectrum disorders,'' {\em Frontiers
  in Robotics and AI}, p.~110, 2019.

\bibitem{baxter2017robot}
P.~Baxter, E.~Ashurst, R.~Read, J.~Kennedy, and T.~Belpaeme, ``Robot education
  peers in a situated primary school study: Personalisation promotes child
  learning,'' {\em PloS one}, vol.~12, no.~5, p.~e0178126, 2017.

\bibitem{scassellati2018improving}
B.~Scassellati {\em et~al.}, ``Improving social skills in children with asd
  using a long-term, in-home social robot,'' {\em Science Robotics}, vol.~3,
  no.~21, 2018.

\bibitem{tsiakas2018task}
K.~Tsiakas, M.~Abujelala, and F.~Makedon, ``Task engagement as personalization
  feedback for socially-assistive robots and cognitive training,'' {\em
  Technologies}, vol.~6, no.~2, p.~49, 2018.

\bibitem{gordon2016affective}
G.~Gordon {\em et~al.}, ``Affective personalization of a social robot tutor for
  children’s second language skills,'' in {\em Proc. AAAI Conf. Artificial
  Intelligence}, vol.~30, 2016.

\bibitem{singh1996reinforcement}
S.~P. Singh and R.~S. Sutton, ``Reinforcement learning with replacing
  eligibility traces,'' {\em Machine learning}, vol.~22, no.~1, pp.~123--158,
  1996.

\bibitem{ortony1990cognitive}
A.~Ortony, G.~L. Clore, and A.~Collins, {\em The Cognitive Structure of
  Emotions}.
\newblock Cambridge University Press, 1990.

\bibitem{martins2017bum}
G.~S. Martins, L.~Santos, and J.~Dias, ``Bum: Bayesian user model for
  distributed social robots,'' in {\em 2017 26th IEEE Int. Symposium on Robot
  and Human Interactive Communication (RO-MAN)}, pp.~1279--1284, IEEE, 2017.

\bibitem{martins2018bum}
G.~S. Martins, L.~Santos, and J.~Dias, ``Bum: Bayesian user model for
  distributed learning of user characteristics from heterogeneous
  information,'' {\em IEEE Transactions on Cognitive and Developmental
  Systems}, vol.~11, no.~3, pp.~425--434, 2018.

\bibitem{martins2019alphapomdp}
G.~S. Martins, H.~Al~Tair, L.~Santos, and J.~Dias, ``$\alpha${POMDP}:
  {POMDP}-based user-adaptive decision-making for social robots,'' {\em Pattern
  Recognition Letters}, vol.~118, pp.~94--103, 2019.

\bibitem{karami2016adaptive}
A.~B. Karami, K.~Sehaba, and B.~Encelle, ``Adaptive artificial companions
  learning from users’ feedback,'' {\em Adaptive Behavior}, vol.~24, no.~2,
  pp.~69--86, 2016.

\bibitem{javed2019robotic}
H.~Javed, R.~Burns, M.~Jeon, A.~M. Howard, and C.~H. Park, ``A robotic
  framework to facilitate sensory experiences for children with autism spectrum
  disorder: A preliminary study,'' {\em ACM Transactions on Human-Robot
  Interaction (THRI)}, vol.~9, no.~1, pp.~1--26, 2019.

\bibitem{tsiakas2016intelligent}
K.~Tsiakas, M.~Abujelala, A.~Lioulemes, and F.~Makedon, ``An intelligent
  interactive learning and adaptation framework for robot-based vocational
  training,'' in {\em 2016 IEEE Symposium Series on Computational Intelligence
  (SSCI)}, pp.~1--6, IEEE, 2016.

\end{thebibliography}

\end{document}